\pdfoutput=1
\documentclass[final,12pt]{clear2024}

\usepackage{times}
\usepackage{tikz}
\usetikzlibrary{arrows,arrows.meta,automata,backgrounds,calc,patterns,positioning,shapes,shadows}
\usepackage[capitalise]{cleveref}
\usepackage{acronym}
\usepackage{booktabs}
\usepackage{csquotes}

\title{Lifted Causal Inference in Relational Domains}

\clearauthor{
	\Name{Malte Luttermann} \Email{malte.luttermann@dfki.de}\\
	\addr German Research Center for Artificial Intelligence (DFKI), Lübeck, Germany
	\AND
	\Name{Mattis Hartwig} \Email{mattis.hartwig@dfki.de}\\
	\addr German Research Center for Artificial Intelligence (DFKI), Lübeck, Germany\\
	singularIT GmbH, Leipzig, Germany
	\AND
	\Name{Tanya Braun} \Email{tanya.braun@uni-muenster.de}\\
	\addr Data Science Group, University of Münster, Germany
	\AND
	\Name{Ralf Möller} \Email{moeller@ifis.uni-luebeck.de}\\
	\addr Institute of Information Systems, University of Lübeck, Germany\\
	German Research Center for Artificial Intelligence (DFKI), Lübeck, Germany
	\AND
	\Name{Marcel Gehrke} \Email{gehrke@ifis.uni-luebeck.de}\\
	\addr Institute of Information Systems, University of Lübeck, Germany
}

\acrodef{ace}[ACE]{average causal effect}
\acrodef{bn}[BN]{Bayesian network}
\acrodef{cp}[CP]{colour passing}
\acrodef{cpr}[ACP]{advanced colour passing}
\acrodef{crv}[CRV]{counting randvar}
\acrodef{fg}[FG]{factor graph}
\acrodef{lci}[LCI]{lifted causal inference}
\acrodef{ljt}[LJT]{lifted junction tree}
\acrodef{lv}[logvar]{logical variable}
\acrodef{lve}[LVE]{lifted variable elimination}
\acrodef{pcfg}[PCFG]{parametric causal factor graph}
\acrodef{pcrv}[PCRV]{parameterised CRV}
\acrodef{pf}[parfactor]{parametric factor}
\acrodef{pfg}[PFG]{parametric factor graph}
\acrodef{prv}[PRV]{parameterised randvar}
\acrodef{rcm}[RCM]{relational causal model}
\acrodef{rv}[randvar]{random variable}
\acrodef{ve}[VE]{variable elimination}
\acrodef{wl}[WL]{Weisfeiler-Leman}

\crefname{algorithm}{Alg.}{Algs.}
\Crefname{algorithm}{Algorithm}{Algorithms}
\crefname{definition}{Def.}{Defs.}

\newcommand{\abs}[1]{\lvert #1 \rvert}

\newcommand{\upmodels}{\ensuremath{\perp\!\!\!\perp}}

\newcommand{\Ch}{\mathrm{Ch}}
\newcommand{\Pa}{\mathrm{Pa}}

\definecolor{myyellow}{RGB}{247,192,26}
\definecolor{myblue}{RGB}{37,122,164}
\definecolor{mygreen}{RGB}{78,155,133}
\definecolor{mypurple}{RGB}{86,51,94}

\pgfdeclarelayer{bg}
\pgfsetlayers{bg,main}

\tikzset{
	rv/.style={draw, ellipse},
	pf/.style={draw, rectangle, fill = gray!30},
	arc/.style = {->, >={[round,sep]Stealth}},
}

\newcommand\factorat[4]{
	\node[pf, label={#2:{#3}}](#4) at (#1) {};
}

\newcommand\factor[6]{
	\node[pf, #1=#3 of #2, label={#4:{#5}}](#6) {};
}

\newcommand\pfs[8]{
	\node[pf, #1=#3 of #2, xshift=-1mm, yshift=1mm](#6) {};
	\node[pf, #1=#3 of #2, label={[label distance=1mm]#4:{#5}}](#7) {};
	\node[pf, #1=#3 of #2, xshift=1mm, yshift=-1mm](#8) {};
}

\newcommand\pfsat[6]{
	\node[pf, xshift=-1mm, yshift=1mm](#4) at (#1) {};
	\node[pf, label={[label distance=1mm]#2:{#3}}](#5) at (#1) {};
	\node[pf, xshift=1mm, yshift=-1mm](#6) at (#1) {};
}

\begin{document}

\maketitle

\begin{abstract}
	Lifted inference exploits symmetries in probabilistic graphical models by using a representative for indistinguishable objects, thereby speeding up query answering while maintaining exact answers.
	Even though lifting is a well-established technique for the task of probabilistic inference in relational domains, it has not yet been applied to the task of causal inference.
	In this paper, we show how lifting can be applied to efficiently compute causal effects in relational domains.
	More specifically, we introduce \aclp{pcfg} as an extension of \aclp{pfg} incorporating causal knowledge and give a formal semantics of interventions therein.
	We further present the \acl{lci} algorithm to compute causal effects on a lifted level, thereby drastically speeding up causal inference compared to propositional inference, e.g., in causal \aclp{bn}.
	In our empirical evaluation, we demonstrate the effectiveness of our approach.
\end{abstract}


\acresetall

\section{Introduction}
A fundamental problem in the research field of artificial intelligence for an intelligent agent is to plan and act rationally in a relational domain.
To compute the best possible action in a perceived state, the agent considers the available actions and chooses the one with the maximum expected utility.
When computing the expected utility of an action that acts on a specific variable, it is crucial to deploy the semantics of an intervention instead of a typical conditioning on that variable~\citep[Chapter~4]{Pearl2009a}.
When calculating the effect of an intervention, a specific variable is set to a fixed value and all incoming probabilistic causal influences of this variable must be ignored for the specific query.
It is therefore fundamental to deploy the semantics of an intervention instead of the typical conditioning to correctly determine the effect of an action.

Over the last years, causal graphical models have become a widely used formalism to answer questions concerning the causal impact of a treatment variable on an outcome variable.
These models combine probabilistic modeling with causal knowledge, which enables computing the effect of an action that intervenes on a particular variable.
As our world is inherently relational (i.e., it consists of objects and relations between those objects), it is particularly important to have models that represent the relational structure between objects in addition to capturing causal knowledge.
However, commonly applied causal graphical models focus on propositional representations while at the same time relational models lack the ability to efficiently apply causal knowledge for inference.
Therefore, we aim to combine the best of both worlds to allow for efficient causal inference in relational domains.
In particular, this paper deals with the problem of efficiently computing causal effects in models representing objects and their causal relationships to each other.

\paragraph{Previous work.}
To perform causal effect estimation in causal graphical models, there has been a considerable amount of work and most of this work focuses on models of propositional data~\citep{Spirtes2000a,Pearl2009a}.
Some works extend propositional \acp{fg} by adding edge directions to enable the computation of the effect of interventions~\citep{Frey2003a,Winn2012a}.
\citet{Maier2010a} show that propositional models are insufficient to represent causal relationships within relational domains as required by real-world applications.
To express causal dependencies within relational domains, \citet{Maier2013a} introduce so-called \acp{rcm} but focus on learning \acp{rcm} from observed data.
Further related work covering \acp{rcm} also focuses on causal discovery and reasoning about conditional independence (e.g., \citealp{Lee2015a,Lee2016a,Lee2019a}).
\Acp{rcm} have also been extended to cover cyclic dependency structures~\citep{Ahsan2022a,Ahsan2023a}, however, both non-cyclic and cyclic \acp{rcm} allow only for reasoning about conditional independence on a lifted level but they do not allow for lifted causal inference.
Prior work dealing with the estimation of causal effects in relational domains applies propositional probabilistic inference~\citep{Arbour2016a,Salimi2020a} and thus does not scale for large graphs.
Consequently, there is a lack of efficient algorithms to compute causal effects in relational domains.
In probabilistic inference, lifting exploits symmetries in a relational model, allowing to carry out query answering more efficiently while maintaining exact answers~\citep{Niepert2014a}.
First introduced by \citet{Poole2003a}, \acp{pfg} and \ac{lve} allow to perform lifted probabilistic inference, i.e., to exploit symmetries in a probabilistic graphical model, resulting in significant speed-ups for probabilistic query answering in relational domains.
Over time, \ac{lve} has been refined by many researchers to reach its current form~\citep{DeSalvoBraz2005a,DeSalvoBraz2006a,Milch2008a,Kisynski2009a,Taghipour2013a,Braun2018a}.
\Acp{pfg} are well-studied for many years and have been developed further to efficiently perform probabilistic inference not only for single queries but also for sets of queries~\citep{Braun2016a}, to incorporate probabilistic inference over time~\citep{Gehrke2018a,Gehrke2020a}, and, among other extensions, to allow for decision making by following the maximum expected utility principle~\citep{Gehrke2019c,Gehrke2019b,Braun2022a}.
Markov logic networks are another lifted representation and have been extended to incorporate maximum expected utility as well~\citep{Apsel2012a}.
Nevertheless, when a decision-making agent plans for the best action to take, previous works improperly apply conditioning (i.e., actions are treated as evidence), as also suggested by \citet[Chapter~16]{Russell2020a}, instead of the notion of an intervention.
Treating actions as evidence, however, is incorrect as noted by \citet[Chapter~4]{Pearl2009a}.
To correctly handle the semantics of an action, the notion of an intervention~\citep{Pearl2016a} has to be applied.
Therefore, in this paper, we close the gap between \acp{pfg} and causal inference in relational domains by introducing \aclp{pcfg} as an extension of \aclp{pfg} incorporating causal knowledge to allow for lifted causal inference, thereby enabling efficient decision making using the notion of an intervention.

\paragraph{Our contributions.}
\Acp{pfg} are well-established models coming with \ac{lve} as a mature lifted inference algorithm, allowing for tractable probabilistic inference with respect to domain sizes in relational domains.
We extend \acp{pfg} by incorporating causal knowledge, resulting in \acp{pcfg} for which we define a formal semantics of interventions.
Having defined a formal semantics of interventions in \acp{pcfg}, we show how causal effects can be efficiently computed, even for multiple simultaneous interventions.
We further introduce the \ac{lci} algorithm that operates on a lifted level to allow for lifted causal inference in relational domains.
Apart from the theoretical investigation of \acp{pcfg} and the \ac{lci} algorithm, we provide an empirical evaluation confirming the efficiency of \ac{lci}.

\paragraph{Structure of this paper.}
In \cref{sec:lce_prelim}, we introduce both \acp{fg} and \acp{pfg} as undirected probabilistic graphical models.
Thereafter, in \cref{sec:lce_pcfg}, we define \acp{pcfg} as an extension of \acp{pfg} incorporating causal knowledge and provide a formal semantics of interventions in \acp{pcfg}.
In \cref{sec:lce_cee}, we introduce the \ac{lci} algorithm operating on a \ac{pcfg} and show how \ac{lci} computes causal effects on a lifted level to avoid grounding the \ac{pcfg} as much as possible.
Afterwards, in our empirical evaluation in \cref{sec:lce_eval}, we investigate the speed-up of \ac{lci} compared to performing propositional causal inference both in causal \aclp{bn} and in directed \acp{fg} before we conclude in \cref{sec:lce_conclusion}.

\section{Preliminaries} \label{sec:lce_prelim}
We begin by introducing \acp{fg} as propositional probabilistic models and afterwards continue to define \acp{pfg} which combine probabilistic models and first-order logic to allow for tractable probabilistic inference with respect to domain sizes in relational domains.
An \ac{fg} is an undirected graphical model to compactly encode a full joint probability distribution between \acp{rv}~\citep{Frey1997a,Kschischang2001a}.
Similar to a \acl{bn}~\citep{Pearl1988a}, an \ac{fg} factorises a full joint probability distribution into a product of factors.
\begin{definition}[Factor Graph]
	An \emph{\ac{fg}} $G = (\boldsymbol V, \boldsymbol E)$ is a bipartite graph with node set $\boldsymbol V = \boldsymbol R \cup \boldsymbol \Phi$ where $\boldsymbol R = \{R_1, \ldots, R_n\}$ is a set of variable nodes (\acp{rv}) and $\boldsymbol \Phi = \{\phi_1, \ldots, \phi_m\}$ is a set of factor nodes (functions).
	There is an edge between a variable node $R_i$ and a factor node $\phi_j$ in $\boldsymbol E \subseteq \boldsymbol R \times \boldsymbol \Phi$ if $R_i$ appears in the argument list of $\phi_j$.
	A factor is a function that maps its arguments to a positive real number, called potential.
	The semantics of $G$ is given by
	$$
		P_G = \frac{1}{Z} \prod_{j=1}^m \phi_j(\mathcal A_j)
	$$
	with $Z$ being the normalisation constant and $\mathcal A_j$ denoting the \acp{rv} appearing in $\phi_j$.
\end{definition}
\begin{figure}[t]
	\centering
	\begin{tikzpicture}
		\node[rv, draw, minimum width = 2.0cm] (rev) {$Rev$};
		\node[rv, draw, above left  = 0.8cm and 0.1cm of rev, minimum width = 3.1cm] (com_b) {$Comp.bob$};
		\node[rv, draw,       left  = 0.7cm of com_b,         minimum width = 3.1cm] (com_a) {$Comp.alice$};
		\node[rv, draw, above right = 0.8cm and 0.1cm of rev, minimum width = 3.1cm] (com_d) {$Comp.dave$};
		\node[rv, draw,       right = 0.7cm of com_d,         minimum width = 3.1cm] (com_e) {$Comp.eve$};
	
		\factorat{$(com_a.east)!0.4!(rev.west)$}{270}{$\phi_{4}^{3}$}{f1}
		\factorat{$(com_b.east)!0.6!(rev.west)$}{[label distance=-1.5mm]45}{$\phi_{4}^{1}$}{f2}
		\factorat{$(com_d.west)!0.6!(rev.east)$}{[label distance=-1.5mm]135}{$\phi_{4}^{2}$}{f3}
		\factorat{$(com_e.west)!0.4!(rev.east)$}{270}{$\phi_{4}^{4}$}{f4}
	
		\node[rv, draw, above = 0.6cm of com_a, minimum width = 3.7cm] (train_a1) {$Train.alice.t_1$};
		\node[rv, draw, below = 1.3cm of com_a, minimum width = 3.7cm] (train_a2) {$Train.alice.t_2$};
		\node[rv, draw, above = 0.6cm of com_b, minimum width = 3.7cm] (train_b1) {$Train.bob.t_1$};
		\node[rv, draw, below = 1.3cm of com_b, minimum width = 3.7cm] (train_b2) {$Train.bob.t_2$};
		\node[rv, draw, above = 0.6cm of com_d, minimum width = 3.7cm] (train_d1) {$Train.dave.t_1$};
		\node[rv, draw, below = 1.3cm of com_d, minimum width = 3.7cm] (train_d2) {$Train.dave.t_2$};
		\node[rv, draw, above = 0.6cm of com_e, minimum width = 3.7cm] (train_e1) {$Train.eve.t_1$};
		\node[rv, draw, below = 1.3cm of com_e, minimum width = 3.7cm] (train_e2) {$Train.eve.t_2$};
	
		\factorat{$(com_a)!0.5!(train_a1)$}{180}{$\phi_{3}^{1}$}{f5}
		\factorat{$(com_a)!0.5!(train_a2)$}{180}{$\phi_{3}^{5}$}{f6}
		\factorat{$(com_b)!0.5!(train_b1)$}{180}{$\phi_{3}^{2}$}{f7}
		\factorat{$(com_b)!0.6!(train_b2)$}{180}{$\phi_{3}^{6}$}{f8}
		\factorat{$(com_d)!0.5!(train_d1)$}{0}{$\phi_{3}^{3}$}{f9}
		\factorat{$(com_d)!0.6!(train_d2)$}{0}{$\phi_{3}^{7}$}{f10}
		\factorat{$(com_e)!0.5!(train_e1)$}{0}{$\phi_{3}^{4}$}{f11}
		\factorat{$(com_e)!0.5!(train_e2)$}{0}{$\phi_{3}^{8}$}{f12}
	
		\node[rv, draw, above = 3.35cm of rev] (qual_t1) {$Qual.t_1$};
		\node[rv, draw, below = 1.3cm of rev] (qual_t2) {$Qual.t_2$};
	
		\factor{below}{qual_t1}{0.1cm}{180}{$\phi_{1}^{1}$}{f13}
		\factor{above}{train_a1}{0.3cm}{180}{$\phi_{2}^{1}$}{f14}
		\factor{above}{train_b1}{0.3cm}{180}{$\phi_{2}^{2}$}{f15}
		\factor{above}{train_d1}{0.3cm}{0}{$\phi_{2}^{3}$}{f16}
		\factor{above}{train_e1}{0.3cm}{0}{$\phi_{2}^{4}$}{f17}
	
		\factor{above}{qual_t2}{0.1cm}{180}{$\phi_{1}^{2}$}{f18}
		\factor{below}{train_a2}{0.3cm}{180}{$\phi_{2}^{5}$}{f19}
		\factor{below}{train_b2}{0.3cm}{180}{$\phi_{2}^{6}$}{f20}
		\factor{below}{train_d2}{0.3cm}{0}{$\phi_{2}^{7}$}{f21}
		\factor{below}{train_e2}{0.3cm}{0}{$\phi_{2}^{8}$}{f22}
	
		\begin{pgfonlayer}{bg}
			\draw (com_a) -- (f1);
			\draw (com_b) -- (f2);
			\draw (com_d) -- (f3);
			\draw (com_e) -- (f4);
			\draw (f1) -- (rev);
			\draw (f2) -- (rev);
			\draw (f3) -- (rev);
			\draw (f4) -- (rev);
			\draw (train_a1) -- (f5);
			\draw (train_a2) -- (f6);
			\draw (train_b1) -- (f7);
			\draw (train_b2) -- (f8);
			\draw (train_d1) -- (f9);
			\draw (train_d2) -- (f10);
			\draw (train_e1) -- (f11);
			\draw (train_e2) -- (f12);
			\draw (f5) -- (com_a);
			\draw (f6) -- (com_a);
			\draw (f7) -- (com_b);
			\draw (f8) -- (com_b);
			\draw (f9) -- (com_d);
			\draw (f10) -- (com_d);
			\draw (f11) -- (com_e);
			\draw (f12) -- (com_e);
			\draw (qual_t1) -- (f13);
			\draw (qual_t1) -- (f14);
			\draw (qual_t1) -- (f15);
			\draw (qual_t1) -- (f16);
			\draw (qual_t1) -- (f17);
			\draw (train_a1) -- (f14);
			\draw (train_b1) -- (f15);
			\draw (train_d1) -- (f16);
			\draw (train_e1) -- (f17);
			\draw (qual_t2) -- (f18);
			\draw (qual_t2) -- (f19);
			\draw (qual_t2) -- (f20);
			\draw (qual_t2) -- (f21);
			\draw (qual_t2) -- (f22);
			\draw (train_a2) -- (f19);
			\draw (train_b2) -- (f20);
			\draw (train_d2) -- (f21);
			\draw (train_e2) -- (f22);
		\end{pgfonlayer}
	\end{tikzpicture}
	\caption{A toy example of an \ac{fg} modelling the interplay of a company's revenue and its employees' competences, which, in turn, can be improved by training employees with a specific training program. We omit the input-output pairs of the factors for brevity.}
	\label{fig:fg_example}
\end{figure}
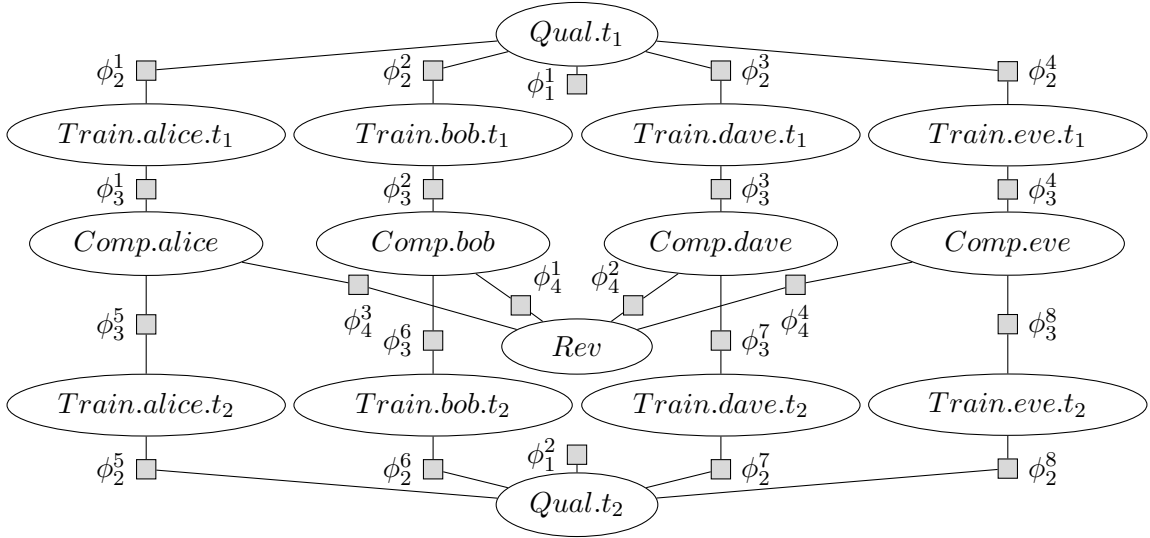
\begin{example} \label{ex:fg_example}
	\Cref{fig:fg_example} shows a toy example of an \ac{fg} modelling the relationships between a company's revenue, the competences of its employees, and the training of its employees.
	More specifically, there are \acp{rv} $Qual.t_i$ indicating the quality of a training program $t_i$, \acp{rv} $Comp.e_j$ describing the competence of an employee $e_j$, \acp{rv} $Train.e_j.t_i$ specifying whether employee $e_j$ has been trained with training program $t_i$, and a \ac{rv} $Rev$ denoting the revenue of the company.
	In this particular example, there is a single company with four employees $alice$, $bob$, $dave$, and $eve$ and there are two training programs $t_1$ and $t_2$ each employee can be trained with.
	The \acp{rv} $Qual.t_i$, $Comp.e_j$, and $Rev$ can take one of the values $\{low,medium,high\}$ and the \acp{rv} $Train.e_j.t_i$ are Boolean.
	Factors $\phi_1^i$ encode the prior probability distribution for the quality of a training program, factors $\phi_2^i$ encode the relationship between the quality of a training program and an employee being trained with that program, factors $\phi_3^i$ encode the relationship between an employee being trained by a specific training program and the competence of that employee, and factors $\phi_4^i$ encode the relationship between the competence of an employee and the revenue of the company.
	The input-output pairs of the factors are omitted for brevity.
\end{example}
We continue to define \acp{pfg}, first introduced by \citet{Poole2003a}, which combine probabilistic models and first-order logic.
In particular, \acp{pfg} use \acp{lv} as parameters in \acp{rv} to represent sets of indistinguishable \acp{rv}.
Each set of indistinguishable \acp{rv} is represented by a so-called \ac{prv}, defined as follows.
\begin{definition}[Parameterised Random Variable]
	Let $\boldsymbol{R}$ be a set of \ac{rv} names, $\boldsymbol{L}$ a set of \ac{lv} names, and $\boldsymbol{D}$ a set of constants.
	All sets are finite.
	Each \ac{lv} $L$ has a domain $\mathcal{D}(L) \subseteq \boldsymbol{D}$.
	A \emph{constraint} is a tuple $(\mathcal{X}, C_{\mathcal{X}})$ of a sequence of \acp{lv} $\mathcal{X} = (X_1, \ldots, X_n)$ and a set $C_{\mathcal{X}} \subseteq \times_{i = 1}^n\mathcal{D}(X_i)$.
	The symbol $\top$ for $C$ marks that no restrictions apply, i.e., $C_{\mathcal{X}} = \times_{i = 1}^n\mathcal{D}(X_i)$.
	A \emph{\ac{prv}} $R(L_1, \ldots, L_n)$, $n \geq 0$, is a syntactical construct of a \ac{rv} $R \in \boldsymbol{R}$ possibly combined with \acp{lv} $L_1, \ldots, L_n \in \boldsymbol{L}$ to represent a set of \acp{rv}.
	If $n = 0$, the \ac{prv} is parameterless and forms a propositional \ac{rv}.
	A \ac{prv} $A$ (or \ac{lv} $L$) under constraint $C$ is given by $A_{|C}$ ($L_{|C}$), respectively.
	We may omit $|\top$ in $A_{|\top}$ or $L_{|\top}$.
	The term $\mathcal{R}(A)$ denotes the possible values (range) of a \ac{prv} $A$. 
	An \emph{event} $A = a$ denotes the occurrence of \ac{prv} $A$ with range value $a \in \mathcal{R}(A)$.
\end{definition}
\begin{example} \label{ex:prv_example}
	Consider $\boldsymbol{R} = \{Qual, Train, Comp, Rev\}$ for quality, training, competence, and revenue, respectively, $\boldsymbol{L} = \{E,T\}$ with $\mathcal{D}(E) = \{alice, bob, dave, eve\}$ (employees) and $\mathcal{D}(T) = \{t_1, t_2\}$ (training programs), combined into \acp{prv} $Qual(T)$, $Train(E,T)$, $Comp(E)$, and $Rev$.
\end{example}
A \ac{pf} describes a function, mapping argument values to positive real numbers, of which at least one is non-zero.
\begin{definition}[Parfactor]
	Let $\Phi$ denote a set of factor names.
	We denote a \emph{\ac{pf}} $g$ by $\phi(\mathcal{A})_{| C}$ with $\mathcal{A} = (A_1, \ldots, A_n)$ being a sequence of \acp{prv}, $\phi$$:$ $\times_{i = 1}^n \mathcal{R}(A_i) \mapsto \mathbb{R}^+$ being a function with name $\phi \in \Phi$ mapping argument values to a positive real number called \emph{potential}, and $C$ being a constraint on the \acp{lv} of $\mathcal{A}$.
	We may omit $|\top$ in $\phi(\mathcal{A})_{|\top}$.
	The term $lv(Y)$ refers to the \acp{lv} in some element $Y$, a \ac{prv}, a \ac{pf}, or sets thereof.
	The term $gr(Y_{| C})$ denotes the set of all instances (groundings) of $Y$ with respect to constraint $C$.
\end{definition}
\begin{example} \label{ex:pf_example}
	Take a look at the \ac{pf} $g_3 = \phi_3(Train(E,T), Comp(E))_{| \top}$.
	Assuming the same ranges of the \acp{prv} and the same domains of the \acp{lv} as in \cref{ex:fg_example,ex:prv_example}, $g_3$ specifies $2 \cdot 3 = 6$ input-output pairs $\phi_3(true, low) = \varphi_1$, $\phi_3(true, medium) = \varphi_2$, $\phi_3(true, high) = \varphi_3$, and so on with $\varphi_i \in \mathbb{R}^+$.
	Further, $lv(g_3) = \{E,T\}$ and $gr(g_{3_{\top}}) = \{\phi_3(Train(alice,t_1), Comp(alice))$, \dots, $\phi_3(Train(eve,t_2), Comp(eve))\}$.
	Thus, in this specific example, the \ac{pf} $g_3$ is able to represent a set of eight factors.
\end{example}
A \ac{pfg} is then built from a set of \acp{pf} $\{g_1, \dots, g_m\}$.
\begin{definition}[Parametric Factor Graph]
	A \emph{\ac{pfg}} $G = (\boldsymbol V, \boldsymbol E)$ is a bipartite graph with node set $\boldsymbol V = \boldsymbol A \cup \boldsymbol G$ where $\boldsymbol A = \{A_1, \ldots, A_n\}$ is a set of \acp{prv} and $\boldsymbol G = \{g_1, \ldots, g_m\}$ is a set of \acp{pf}.
	A \ac{prv} $A_i$ and a \ac{pf} $g_j$ are connected via an edge in $G$ (i.e., $\{A_i, g_j\} \in \boldsymbol E$) if $A_i$ appears in the argument list of $g_j$.
	The semantics of $G$ is given by grounding and building a full joint distribution.
	With $Z$ as the normalisation constant and $\mathcal A_k$ denoting the \acp{rv} connected to $\phi_k$, $G$ represents the full joint distribution
	$$
		P_G = \frac{1}{Z} \prod_{g_j \in \boldsymbol G} \prod_{\phi_k \in gr(g_j)} \phi_k(\mathcal A_k).
	$$
\end{definition}
\begin{figure}[t]
	\centering
	\begin{tikzpicture}
		\node[rv, draw, inner sep = 2.0pt] (qual) {$Qual(T)$};
		\node[rv, draw, right = 1.4cm of qual,  inner sep = 2.0pt] (train) {$Train(E,T)$};
		\node[rv, draw, right = 1.4cm of train, inner sep = 2.0pt] (comp) {$Comp(E)$};
		\node[rv, draw, right = 1.4cm of comp,  minimum width = 1.8cm] (rev) {$Rev$};
	
		\pfs{left}{qual}{0.8cm}{270}{$g_1$}{f1a}{f1}{f1b}
		\pfsat{$(qual.east)!0.5!(train.west)$}{270}{$g_2$}{f2a}{f2}{f2b}
		\pfsat{$(train.east)!0.5!(comp.west)$}{270}{$g_3$}{f3a}{f3}{f3b}
		\pfsat{$(comp.east)!0.5!(rev.west)$}{270}{$g_4$}{f4a}{f4}{f4b} 
	
		\begin{pgfonlayer}{bg}
			\draw (f1) -- (qual);
			\draw (qual) -- (f2);
			\draw (f2) -- (train);
			\draw (train) -- (f3);
			\draw (f3) -- (comp);
			\draw (comp) -- (f4);
			\draw (f4) -- (rev);
		\end{pgfonlayer}
	\end{tikzpicture}
	\caption{A visualisation of a \ac{pfg} entailing equivalent semantics as the \ac{fg} shown in \cref{fig:fg_example}. Each \ac{pf} represents a group of factors and each \ac{prv} represents a group of \acp{rv}.}
	\label{fig:pfg_example}
\end{figure}
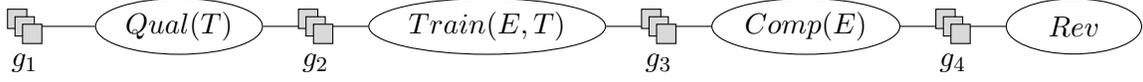
\begin{example}
	\Cref{fig:pfg_example} depicts a \ac{pfg} $G$ consisting of four \acp{pf} $g_1 = \phi_1(Qual(T))$, $g_2 = \phi_2(Qual(T), Train(E,T))$, $g_3 = \phi_3(Train(E,T), Comp(E))$, and $g_4 = \phi_4(Comp(E), Rev)$.
	Assuming that both the ranges of the \acp{prv} and the domains of the \acp{lv} follow \cref{ex:fg_example,ex:prv_example,ex:pf_example}, $G$ is a lifted representation entailing equivalent semantics as the \ac{fg} shown in \cref{fig:fg_example}.
	Each \ac{pf} $g_1$, \dots, $g_4$ represents a group of factors $\phi_1^i$, \dots, $\phi_4^i$, respectively, and each \ac{prv} $Qual$, $Train$, $Comp$, and $Rev$ represents a group of \acp{rv}.
\end{example}
The underlying assumption here is that there are indistinguishable objects, in this specific example employees, which can be represented by a representative.
In particular, the assumption is that the competence of every employee has the same influence on the company's revenue, i.e., all factors $\phi_4^i$ encode the same mappings (and the same holds for the factors $\phi_1^i$, $\phi_2^i$, and $\phi_3^i$, meaning training programs are indistinguishable as well).
In other words, it is relevant for the company how many employees are competent but it does not matter which exact employees are competent.
Note that the definition of \acp{pfg} also includes \acp{fg}, as every \ac{fg} is a \ac{pfg} containing only parameterless \acp{rv}.

In the following, we extend \acp{pfg} to incorporate causal knowledge, represented by directed edges defining cause-effect relationships between \acp{prv}.

\section{Parametric Causal Factor Graphs} \label{sec:lce_pcfg}
\Acp{pfg} are well-established models for which lifted inference algorithms exist to allow for tractable probabilistic inference with respect to domain sizes.
Even though \citet{Frey2003a} introduces directed \acp{fg} to allow for representing causal knowledge on a ground level, \acp{pfg} have not yet been extended to incorporate causal knowledge on a lifted level.

\begin{figure}[t]
	\centering
	\begin{tikzpicture}
		\node[rv, draw, inner sep = 2.0pt] (qual) {$Qual(T)$};
		\node[rv, draw, right = 1.4cm of qual,  inner sep = 2.0pt] (train) {$Train(E,T)$};
		\node[rv, draw, right = 1.4cm of train, inner sep = 2.0pt] (comp) {$Comp(E)$};
		\node[rv, draw, right = 1.4cm of comp,  minimum width = 1.8cm] (rev) {$Rev$};
	
		\pfs{left}{qual}{0.8cm}{270}{$g_1$}{f1a}{f1}{f1b}
		\pfsat{$(qual.east)!0.5!(train.west)$}{270}{$g_2$}{f2a}{f2}{f2b}
		\pfsat{$(train.east)!0.5!(comp.west)$}{270}{$g_3$}{f3a}{f3}{f3b}
		\pfsat{$(comp.east)!0.5!(rev.west)$}{270}{$g_4$}{f4a}{f4}{f4b}
	
		\begin{pgfonlayer}{bg}
			\draw[arc] (f1) -- (qual);
			\draw (qual) -- (f2);
			\draw[arc] (f2) -- (train);
			\draw (train) -- (f3);
			\draw[arc] (f3) -- (comp);
			\draw (comp) -- (f4);
			\draw[arc] (f4) -- (rev);
		\end{pgfonlayer}
	\end{tikzpicture}
	\caption{An illustration of a \ac{pcfg} encoding the same full joint probability distribution as the \ac{pfg} given in \cref{fig:pfg_example}. The only difference between the \ac{pcfg} and the \ac{pfg} is that the \ac{pcfg} contains directed edges instead of undirected edges between \acp{prv} and \acp{pf}.}
	\label{fig:pcfg_example}
\end{figure}
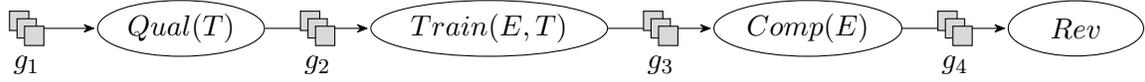

Therefore, we now introduce \acp{pcfg} as an extension of \acp{pfg} incorporating causal knowledge and give a formal semantics of interventions therein.
A \ac{pcfg} extends a \ac{pfg} by incorporating causal knowledge in form of directed edges---that is, all edges between two \acp{prv} (via a \ac{pf}) are directed and describe a cause-effect relationship.
For example, an edge $A_1 \to A_2$ indicates that $A_1$ is a cause of $A_2$ and, consequently, $A_2$ is an effect of $A_1$.
In particular, in a \ac{pcfg}, each \ac{pf} is connected to a single child and zero or more parents, matching the definition of directed \acp{fg} in the ground case given by \citet{Frey2003a}. 
Further, as commonly required in directed graphical models such as causal \aclp{bn}, we restrict a \ac{pcfg} to be acyclic.
\begin{definition}[Parametric Causal Factor Graph] \label{def:pcfg}
	A \emph{\ac{pcfg}} is a fully directed graph $G = (\boldsymbol V, \boldsymbol E)$ with node set $\boldsymbol V = \boldsymbol A \cup \boldsymbol G$ where $\boldsymbol A = \{A_1, \ldots, A_n\}$ is a set of \acp{prv} and $\boldsymbol G = \{g_1, \ldots, g_m\}$ is a set of \emph{directed \acp{pf}}.
	A directed \ac{pf} $g = \phi(\mathcal A)_{| C}^{\rightarrow A_i}$ with $\mathcal A = (A_1, \ldots, A_k)$ being a sequence of \acp{prv}, $\phi$$:$ $\times_{i = 1}^k \mathcal{R}(A_i) \mapsto \mathbb{R}^+$ being a function, and $C$ being a constraint on the \acp{lv} of $\mathcal{A}$, maps its argument values to a positive real number (potential).
	Again, we may omit $|\top$ in $\phi(\mathcal{A})_{|\top}^{\rightarrow A_i}$.
	$A_i \in \mathcal A$ denotes the \emph{child} of $\phi(\mathcal A)^{\rightarrow A_i}$ whereas all $A_j \in \mathcal A$ with $j \neq i$ are the \emph{parents} of $\phi(\mathcal A)^{\rightarrow A_i}$.
	For each directed \ac{pf} $g$, there are edges $(g, A_i) \in \boldsymbol E$ and $(A_j, g) \in \boldsymbol E$ (for all $A_j \neq A_i$).
	A \ac{pcfg} is an acyclic graph, that is, there is no sequence of edges $(g_1, A_1), (A_1, g_2), \dots, (g_{\ell}, A_{\ell}), (A_{\ell}, g_1)$ in $\boldsymbol E$.
	The semantics of $G$ is given by grounding and building a full joint distribution, identical to the semantics of a \ac{pfg}, i.e., with $Z$ as the normalisation constant, $\mathcal A_k$ denoting the \acp{rv} connected to $\phi_k$, and $A_i^k \in \mathcal A_k$ specifying the child of $\phi_k$, $G$ represents
	$$
		P_G = \frac{1}{Z} \prod_{g_j \in \boldsymbol G} \prod_{\phi_k \in gr(g_j)} \phi_k(\mathcal A_k)^{\rightarrow A_i^k}.
	$$
\end{definition}
\begin{example}
	Consider the \ac{pcfg} $G$ depicted in \cref{fig:pcfg_example}.
	$G$ represents the same full joint probability distribution as the \ac{pfg} shown in \cref{fig:pfg_example}.
	In particular, both models are identical except for the fact that $G$ contains directed edges instead of undirected edges between \acp{pf} and \acp{prv}.
	Each \ac{pf} represents a group of directed factors and thus, grounding $G$ results in a directed \ac{fg}.
	Following previous examples by assuming $\mathcal D(T) = \{t_1, t_2\}$, for example $g_1 = \phi_1(Qual(T))^{\rightarrow Qual(T)}$ represents $gr(g_1) = \{\phi_1(Qual(t_1))^{\rightarrow Qual(t_1)}, \allowbreak \phi_1(Qual(t_2))^{\rightarrow Qual(t_2)}\}$.
\end{example}
In the following, we denote the parents of a \ac{prv} $A$ by $\Pa_G(A) = \{\phi^{\rightarrow A_i} \mid A = A_i \}$ and the child of a \ac{pf} $\phi$ by $\Ch_G(\phi(\mathcal A)^{\rightarrow A_i}) = A_i$ in a \ac{pcfg} $G$.
If the context is clear, we omit the subscript $G$.
Before we define the semantics of an intervention in a \ac{pcfg}, we briefly revisit the notion of $d$-separation in directed acyclic graphs and afterwards apply it to \acp{pcfg}.

\subsection{\texorpdfstring{$\boldsymbol d$}{d}-Separation in Parametric Causal Factor Graphs}
The notion of $d$-separation~\citep{Pearl1986a} provides a graphical criterion to test for conditional independence in directed acyclic graphs.
\citet{Frey2003a} translates the notion of $d$-separation to directed \acp{fg}.
We build on the definition of $d$-separation in directed \acp{fg} provided by \citet{Frey2003a} to define $d$-separation in \acp{pcfg} on a ground level.

\begin{definition}[$\boldsymbol d$-separation]
	Let $G = (\boldsymbol A \cup \boldsymbol G, \boldsymbol E)$ be a \ac{pcfg}.
	Given three disjoint sets of \acp{rv} $\boldsymbol X$, $\boldsymbol Y$, and $\boldsymbol Z$ (subsets of $\bigcup_{A \in \boldsymbol A} gr(A)$), we say that $\boldsymbol X$ and $\boldsymbol Y$ are conditionally independent given $\boldsymbol Z$, written as $\boldsymbol X \upmodels \boldsymbol Y \mid \boldsymbol Z$, if the nodes in $\boldsymbol Z$ block all paths from the nodes in $\boldsymbol X$ to the nodes in $\boldsymbol Y$ in the directed \ac{fg} obtained by grounding $G$. 
	A path is a connected sequence of edges $(R_1, g_1)$, \dots, $(R_{\ell}, g_{\ell})$ and is not restricted to follow the arrow directions of the edges.
	Note that it is therefore also possible for a path to pass from a parent of a factor to another parent of the factor.
	A path is blocked by the nodes in $\boldsymbol Z$ if
	\begin{enumerate}
		\item the path contains a variable from $\boldsymbol Z$, or
		\item the path passes from a parent of a directed factor $\phi$ to another parent of $\phi$, and neither the child of $\phi$ nor any of its descendants are in $\boldsymbol Z$.
	\end{enumerate}
\end{definition}
The semantics of $d$-separation in \acp{pcfg} is defined on a ground level.
However, it is possible to check for $d$-separation on a lifted level without having to ground the \ac{pcfg}.
We give a short motivational example to illustrate this idea and leave the algorithmic details for future work. 
\begin{example}
	Consider again the \ac{pcfg} depicted in \cref{fig:pcfg_example} and assume we want to check whether $Qual(t_1) \upmodels Comp(bob) \mid Train(bob, t_1)$ holds.
	In this case, we have assigned $T = t_1$ and $E = bob$, so we only need to examine paths involving this particular assignment of \acp{lv}.
	More specifically, all \acp{prv} on the path with overlapping \acp{lv}, i.e., all \acp{prv} having $T$ or $E$ as a \ac{lv}, are bound to the same assignment.
	Therefore, all paths from $Qual(t_1)$ to $Comp(bob)$ pass through $Train(bob, t_1)$, meaning the conditional independence statement in question holds.
	Note that if there were other paths involving \acp{prv} with non-overlapping \acp{lv}, i.e., \acp{lv} not involved in the sets $\boldsymbol X$ and $\boldsymbol Y$, all \acp{rv} represented by those \acp{prv} need to be in $\boldsymbol Z$ to block those paths.
\end{example}
We close this subsection with a final remark on $d$-separation in \acp{pcfg}.
In contrast to \aclp{bn}, the conditional independence statements induced by a \ac{pcfg} $G$ are not guaranteed to be compatible with arbitrary potential mappings of the directed \acp{pf} in $G$ (i.e., it is not possible to specify the edge directions and arbitrary potential mappings independent of each other).
In other words, the conditional independence statements induced by the graph structure of $G$ must be encoded in the potential mappings of the directed \acp{pf} in $G$.
For example, if the graph structure of $G$ induces a conditional independence statement $X \upmodels Y \mid Z$ for the \acp{rv} $X$, $Y$, and $Z$, the potential mappings of the directed \acp{pf} in $G$ must be specified such that $P(X, Y \mid Z) = P(X \mid Z) \cdot P(Y \mid Z)$ holds\footnote{In a \acl{bn}, the numbers in all conditional probability tables can be specified arbitrarily (except for the only restriction that each row must sum to one) because the structure of the conditional probability tables enforces that all conditional independence statements induced by the \acl{bn}'s structure are also encoded in the numbers, i.e., $P(X, Y \mid Z) = P(X \mid Z) \cdot P(Y \mid Z)$ holds if and only if $X \upmodels Y \mid Z$.}.
To ensure that the conditional independence statements induced by the graph structure of $G$ are compatible with the potential mappings of the directed \acp{pf} in $G$, it is sufficient to specify the potential mappings for each directed \ac{pf} $\phi(A_1, \dots, A_k)^{\to A_i}$ such that all assignments $(a_1, \dots, a_k)$ which differ only at position $i$ sum up to one (which is equivalent to each row in a conditional probability table in a \acl{bn} summing up to one).
Specifying the potential mappings in this way ensures that each directed \ac{pf} $\phi(A_1, \dots, A_k)^{\to A_i}$ encodes a conditional probability distribution that corresponds to exactly one conditional probability distribution in an equivalent \acl{bn}.
Consequently, the conditional independence statements induced by the graph structure of $G$ are compatible with the potential mappings of the directed \acp{pf} in $G$ because there exists a \acl{bn} that induces the same conditional independence statements as $G$ and encodes the same underlying full joint probability distribution as $G$.
Note that this issue arises only when specifying the model by hand, i.e., in case the model is learned from data, the conditional independence statements induced by the data are automatically reflected both in the graph structure and the potential values.

The concept of $d$-separation is important for the computation of the effect of an intervention in the sense that all non-causal paths, so-called backdoor paths, need to be blocked to remove spurious effects.
We next show how these backdoor paths are blocked when performing an intervention and give a formal semantics of an intervention in a \ac{pcfg}.

\subsection{Semantics of Interventions in Parametric Causal Factor Graphs}
To correctly handle the semantics of an action, for example in the setting of a decision-making agent planning for the best action to take, we have to differentiate between \emph{seeing} (conditioning) and \emph{doing} (intervention).
Let us take a look at the \ac{pcfg} shown in \cref{fig:pcfg_example} again.
If we observe (see) an event $Train(bob,t_1) = true$, our belief about the probability distribution of $Qual(t_1)$ might change.
More specifically, the probability of $t_1$ having a high quality might be higher when observing $Train(bob,t_1) = true$ than without the observation under the assumption that the probability of training an employee increases if the quality of a training program is high.
However, if we are interested in the effect an action setting $Train(bob,t_1)$ to $true$, denoted as $do(Train(bob,t_1) = true)$, has on the remaining \acp{prv}, we have to ensure that the belief about the probability of $t_1$ having a high quality remains unchanged as the action itself has no influence on the probability distribution of $Qual(t_1)$.
Therefore, it is crucial to avoid the propagation of information against the edge directions whenever we are interested in the effect of an action.
That is, if we are interested in the effect a specific \ac{rv} $R'$ has on another \ac{rv} $R$, all so-called backdoor paths from $R$ to $R'$ must be blocked.
A backdoor path is a non-causal path, i.e., a backdoor path from $R$ to $R'$ is a path that remains after removing all outgoing edges of $R$.

To account for backdoor paths and correctly handle the semantics of an action, we employ the notion of an intervention.
We first define the notion of an intervention on \acp{rv} and later extend it to allow for $do$-expressions on \acp{prv}. 
An intervention on a \ac{rv} $R$, denoted as $do(R = r)$ with $r \in \mathcal R(R)$, changes the structure of a \ac{pcfg} by removing all parent edges of $R$ and setting $R$ to the value $r$.
By removing the parent edges, all backdoor paths are removed.
Formally, the semantics of an intervention in a \ac{pcfg} is defined as below, following the definition of an intervention in \aclp{bn} provided by \citet{Pearl2016a}.
\begin{definition}[Intervention] \label{def:intervention}
	Let $\boldsymbol R = \{R_1, \dots, R_n\}$ be the set of \acp{rv} obtained by grounding a \ac{pcfg} $G = (\boldsymbol A \cup \boldsymbol G, \boldsymbol E)$, i.e., $\boldsymbol R = \bigcup_{A \in \boldsymbol A} gr(A)$.
	An intervention $do(R_1 = r_1, \dots, R_k = r_k)$ changes the underlying probability distribution such that each factor $\phi(R'_1, \dots, R'_i, \dots, R'_{\ell})^{\rightarrow R'_i}$ with $R'_i \in \{R_1, \dots, R_k\}$ is replaced by a factor $\phi'(R'_1, \dots, R'_i, \dots R'_{\ell})^{\rightarrow R'_i}$ with
	$$
		\phi'(R'_1 = r'_1, \dots, R'_i = r'_i, \dots, R'_{\ell} = r'_{\ell})^{\rightarrow R'_i} =
		\begin{cases}
			1 & \text{if } r_i = r'_i \\
			0 & \text{if } r_i \neq r'_i.
		\end{cases}
	$$
	All $\phi(R'_1, \dots, R'_i, \dots, R'_{\ell})^{\rightarrow R'_i}$ with $R'_i \notin \{R_1, \dots, R_k\}$ remain unchanged.
\end{definition}
By fixing the values of all parent factors, all parent influences are (virtually) removed from the model and hence initial backdoor paths are (virtually) removed from the model as well.
Having defined the semantics of an intervention in a \ac{pcfg}, we are now interested in efficiently computing interventional distributions, i.e., the result of queries that contain $do$-expressions.
Note that in a \ac{pcfg}, all causal effects are identifiable per definition and thus, we do not have to rewrite a query containing $do$-expressions according to the $do$-calculus~\citep{Pearl1995a} to obtain an equivalent query free of $do$-expressions.
In particular, we do not estimate causal effects from observed data but instead compute them in a fully specified model as every \ac{pcfg} encodes a full joint probability distribution which we can modify according to the definition of an intervention and afterwards query the modified distribution to answer any query containing $do$-expressions. 

We next introduce the \ac{lci} algorithm, which handles interventions in \acp{pcfg} efficiently by directly applying the semantics of an intervention on a lifted level.

\section{Efficient Causal Effect Computation in Parametric Causal Factor Graphs} \label{sec:lce_cee}
Now that we have introduced \acp{pcfg}, we study the problem of efficiently computing the effect of interventions in \acp{pcfg}.
A major advantage of using \acp{pcfg} instead of propositional models such as causal \aclp{bn} is that we mostly do not have to fully ground the model to compute the effect of interventions.
Consider again the \ac{pcfg} illustrated in \cref{fig:pcfg_example} and assume we want to compute the interventional distribution $P(Rev \mid do(Train(bob,t_1) = true))$ in $G$.
Note that when intervening on a \ac{rv}, we have to treat it differently than other \acp{rv} in the same group on which we do not intervene.
An intervention $do(Treat(bob,t_1) = true)$ sets the value of $Treat(bob,t_1)$ to $true$ and thus, we have to treat $bob$ differently from $alice$, $dave$, and $eve$---in other words, not all employees are indistinguishable anymore.
Nevertheless, and this is the crucial point, we can still treat $alice$, $dave$, and $eve$ as indistinguishable when computing the interventional distribution.

\begin{algorithm2e}[t]
	\caption{Lifted Causal Inference}
	\label{alg:lcee}
	\DontPrintSemicolon
	\LinesNumbered
	\SetKwInOut{Input}{Input}
	\SetKwInOut{Output}{Output}
	\Input{A \ac{pcfg} $G = (\boldsymbol A \cup \boldsymbol G, \boldsymbol E)$, and a query $P(R_1, \dots, R_{\ell} \mid do(R'_1 = r'_1, \dots, R'_k = r'_k))$ with $\{R_1, \dots, R_{\ell}\} \subseteq \bigcup_{A \in \boldsymbol A} gr(A)$ and $\{R'_1, \dots, R'_k\} \subseteq \bigcup_{A \in \boldsymbol A} gr(A)$.}
	\Output{The interventional distribution $P(R_1, \dots, R_{\ell} \mid do(R'_1 = r'_1, \dots, R'_k = r'_k))$.}
	\BlankLine
	$G' \gets$ Split \acp{pf} in $G$ based on each $R'_i \in \{R'_1, \dots, R'_k\}$\;
	\ForEach{$R'_i \in \{R'_1, \dots, R'_k\}$}{ \label{line:outer_loop}
		\ForEach{$\phi(A_1, \dots, A_z)^{\rightarrow R'_i} \in \Pa_{G'}(R'_i)$}{ \label{line:middle_loop}
			\ForEach{assignment $(a_1, \dots, a_z) \in \mathcal R(A_1) \times \dots \times \mathcal R(A_z)$}{ \label{line:inner_loop}
				Set $\phi(a_1, \dots, a_z) = \begin{cases} 1 & \text{if $(a_1, \dots, a_z)$ assigns $R'_i = r'_i$} \\ 0 & \text{if $(a_1, \dots, a_z)$ assigns $R'_i \neq r'_i$} \end{cases}$
			}
		}
	}
	$P \gets$ Call \ac{lve} to compute $P(R_1, \dots, R_{\ell})$ in $G'$\;
	\Return{$P$}\;
\end{algorithm2e}

\subsection{The Lifted Causal Inference Algorithm}
We now introduce the \ac{lci} algorithm to compute the interventional distribution $P(R_1, \dots, R_{\ell} \mid do(R'_1 = r'_1, \dots, R'_k = r'_k))$ in a \ac{pcfg} $G$.
The entire \ac{lci} algorithm is shown in \cref{alg:lcee}.

First, \ac{lci} splits the \acp{pf} in $G$ based on the intervention variables $R'_i \in \{R'_1, \dots, R'_k\}$.
In particular, splitting \acp{pf} in $G$ results in a modified \ac{pcfg} $G'$ entailing equivalent semantics as $G$~\citep{DeSalvoBraz2005a}.
The procedure of splitting a \ac{pf} works as follows.
Recall that $R'_i = A(L_1 = l_1, \dots, L_j = l_j)$, $l_1 \in \mathcal D(L_1), \dots, l_j \in \mathcal D(L_j)$, is a particular instance of a \ac{prv} $A(L_1, \dots, L_j)$, that is, it holds that $R'_i \in gr(A)$.
The idea behind the splitting procedure is that we would like to separate $gr(A)$ into two sets $gr(A) \setminus \{R'_i\}$ and $\{R'_i\}$, as $R'_i$ has to be treated differently than the remaining instances of $A$.
Therefore, every \ac{pf} $g$ for which there is an instance $\phi \in gr(g)$ such that $R'_i$ appears in the argument list of $\phi$ is split.
Formally, splitting a \ac{pf} $g$ replaces $g$ by two \acp{pf} $g'_{| C'}$ and $g''_{| C''}$ and adapts the constraints of $g'_{| C'}$ and $g''_{| C''}$.
The constraints $C'$ and $C''$ are altered such that the inputs of $g'_{| C'}$ are restricted to all sequences that contain $R'_i$ and the inputs of $g''_{| C''}$ are restricted to the remaining input sequences.
After the splitting procedure, the semantics of the model remains unchanged as the groundings of $G'$ are still the same as the groundings of the initial model $G$---they are just arranged differently across the sets of ground instances.
Having completed the split of all respective \acp{pf}, \ac{lci} next modifies the parents of $R'_i$, i.e., the underlying probability distribution encoded by $G'$ is modified according to the semantics of the intervention $do(R'_i = r'_i)$ with $r'_i \in \mathcal R(R'_i)$.
More specifically, as $R'_i$ is fixed on $r'_i$, all parents $\phi \in \Pa_{G'}(R'_i)$ of $R'_i$ are altered such that all input sequences assigning $R'_i = r'_i$ map to the potential value one while all other input sequences map to zero.
Finally, \ac{lci} computes the result for $P(R_1, \dots, R_{\ell})$ in the modified model $G'$, which is equivalent to the result for $P(R_1, \dots, R_{\ell} \mid do(R'_1 = r'_1, \dots, R'_k = r'_k))$ in the original model $G$.
To perform query answering in $G'$, \ac{lve} can be applied to $G'$ by simply ignoring the edge directions as the semantics of a \ac{pfg} and a \ac{pcfg} are defined identically.

Before we continue to examine the correctness of \cref{alg:lcee}, we take a look at an example.
\begin{figure}[t]
	\centering
	\begin{tikzpicture}
		\node[rv, draw, minimum width = 3.5cm, inner sep = 2.0pt] (train) {$Train(E,T)$};
		\node[rv, draw, above = 0.8cm of train, inner sep = 2.0pt] (train_split) {$Train(bob, t_1)$};
	
		\node[rv, draw, left = 1.3cm of $(train.west)!0.5!(train_split.west)$, inner sep = 2.0pt] (qual) {$Qual(T)$};
		\node[rv, draw, right = 1.3cm of $(train.east)!0.5!(train_split.east)$, inner sep = 2.0pt] (comp) {$Comp(E)$};
		\node[rv, draw, right = 1.3cm of comp,  minimum width = 1.8cm] (rev) {$Rev$};	
	
		\pfs{left}{qual}{0.6cm}{270}{$g_{1_{| \top}}$}{f1a}{f1}{f1b}
		\pfsat{$(qual.east)!0.5!(train.west)$}{270}{$g_{2_{| C_2}}$}{f2a}{f2}{f2b}
		\pfsat{$(qual.east)!0.5!(train_split.west)$}{90}{$g'_{2_{| C'_2}}$}{f2a_split}{f2_split}{f2b_split}
		\pfsat{$(train.east)!0.5!(comp.west)$}{270}{$g_{3_{| C_3}}$}{f3a}{f3}{f3b}
		\pfsat{$(train_split.east)!0.5!(comp.west)$}{90}{$g'_{3_{| C'_3}}$}{f3a_split}{f3_split}{f3b_split}
		\pfsat{$(comp.east)!0.5!(rev.west)$}{270}{$g_{4_{| \top}}$}{f4a}{f4}{f4b}
	
		\begin{pgfonlayer}{bg}
			\draw[arc] (f1) -- (qual);
			\draw (qual.east) -- (f2);
			\draw[arc] (f2) -- (train.west);
			\draw (train.east) -- (f3);
			\draw[arc] (f3) -- (comp.west);
			\draw (comp.east) -- (f4);
			\draw[arc] (f4) -- (rev.west);
	
			\draw (qual.east) -- (f2_split);
			\draw[arc] (f2_split) -- (train_split.west);
			\draw (train_split.east) -- (f3_split);
			\draw[arc] (f3_split) -- (comp.west);
		\end{pgfonlayer}
	\end{tikzpicture}
	\caption{A visualisation of the modified \ac{pcfg} obtained after altering the \ac{pcfg} shown in \cref{fig:pcfg_example} by splitting $g_2$ and $g_3$ to separate $Train(bob,t_1)$ from $Train(E,T)$. Here, the constraints $C_2$ as well as $C_3$ include all instances of $Train(E,T)$ except for $Train(bob,t_1)$ and $C'_2$ as well as $C'_3$ are restricted to the single instance $Train(bob,t_1)$ of $Train(E,T)$. Note that the graph size remains significantly smaller than for the fully grounded model.}
	\label{fig:lifted_do_example}
\end{figure}
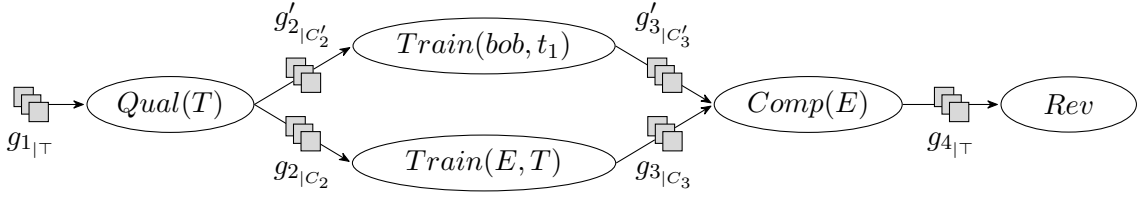
\begin{example}
	Consider again the \ac{pcfg} $G$ shown in \cref{fig:pcfg_example} and assume we would like to compute $P(Rev \mid do(Train(bob,t_1)) = true)$.
	As $Train(bob,t_1)$ is a particular instance of $Train(E,T)$, we have to split the \acp{pf} $g_2$ and $g_3$ while $g_1$ as well as $g_4$ keep $\top$ as their constraint.
	\Cref{fig:lifted_do_example} shows the modified \ac{pcfg} $G'$ obtained after splitting $g_2$ and $g_3$ based on the intervention on $Train(bob,t_1)$.
	In $G'$, $Train(bob,t_1)$ is now a separate node in the graph, connected to two newly introduced \acp{pf}.
	In particular, $g_2$ has been replaced by two \acp{pf} $g_{2_{| C_2}}$ and $g'_{2_{| C'_2}}$ with constraints $C_2 = ((E,T), \{(alice,t_1), \allowbreak (alice,t_2), \allowbreak (bob,t_2), \allowbreak (dave,t_1), \allowbreak (dave,t_2), \allowbreak (eve,t_1), \allowbreak (eve,t_2)\})$ and $C'_2 = ((E,T), \{(bob,t_1)\})$.
	In other words, $g_{2_{| C_2}}$ is restricted to all instances of $Train(E,T)$ except for $Train(bob,t_1)$ and $g'_{2_{| C'_2}}$ is restricted to the instance $Train(bob,t_1)$ of $Train(E,T)$.
	Analogously, $g_3$ has been replaced by two \acp{pf} $g_{3_{| C_3}}$ and $g'_{3_{| C'_3}}$.
	To incorporate the semantics of $do(Train(bob,t_1)) = true$, \ac{lci} next modifies the parents of $Train(bob,t_1)$, i.e., \ac{lci} modifies $g'_{2_{| C'_2}}$ in this example.
	More specifically, $g'_{2_{| C'_2}}(Qual(t_1) = q, Train(bob,t_1) = true)$ is set to one and $g'_{2_{| C'_2}}(Qual(t_1) = q, Train(bob,t_1) = false)$ is set to zero for all $q \in \mathcal R(Qual(t_1))$.
	Finally, \ac{lve} can be run to compute $P(Rev)$ in $G'$, which is equivalent to the interventional distribution $P(Rev \mid do(Train(bob,t_1)) = true)$ in the original model $G$.
\end{example}
Due to the splitting of \acp{pf}, it might be the case that there are \acp{prv} in $G'$ having more parents than they previously had in the original model $G$, as with $Comp(E)$ in \cref{fig:lifted_do_example}.
The semantics of the model, however, remains unchanged because $\bigcup_{g \in \boldsymbol G} gr(g) = \bigcup_{g \in \boldsymbol G'} gr(g)$.
Given the way we specified the semantics of an intervention in a \ac{pcfg}, it immediately follows that \ac{lci} correctly computes the effect of interventions.
In particular, as \ac{lci} directly applies \cref{def:intervention} by setting the parent factors of all variables we intervene on accordingly, the semantics of the modified model $G'$ is equivalent to the semantics of interventions from \cref{def:intervention} and thus, \ac{lci} is correct.
\begin{corollary}
	\Cref{alg:lcee} computes the interventional distribution according to \cref{def:intervention}.
\end{corollary}
Moreover, directly applying \cref{def:intervention} allows \ac{lci} to exploit the established \ac{lve} algorithm.
By deploying \ac{lve}, \ac{lci} is able to perform tractable inference with respect to domain sizes for all \acp{pcfg} in the class of domain-liftable models~\citep{Taghipour2013b} if the $do$-expressions in the query do not ground a domain.
The class of domain-liftable models includes all \acp{pcfg} containing only \acp{pf} with at most two \acp{lv} and all \acp{pcfg} containing only \acp{prv} having at most one \ac{lv}.
\begin{corollary}
	\Cref{alg:lcee} performs tractable probabilistic inference with respect to domain sizes for the class of domain-liftable models if the $do$-expressions in the query do not ground a domain.
\end{corollary}
A more fine-grained complexity analysis of \ac{lve} is provided in \citep{Taghipour2013c}.
Note that the loops in \cref{line:outer_loop,line:middle_loop,line:inner_loop} of \cref{alg:lcee} do not influence the overall run time complexity of \ac{lci} as they iterate over potential mappings that must be considered anyway during inference (both by \ac{lve} and its propositional counterpart \acl{ve}).

To summarise, \ac{lci} is a simple, yet effective algorithm to perform lifted causal inference, even for interventions on large groups of \acp{rv}, as we investigate next.

\subsection{Handling Interventions on Groups of Random Variables} 
\Ac{lci} is able to handle both interventions on a single (ground) \ac{rv} as well as interventions on a conjunction of multiple \acp{rv} efficiently.
In particular, when intervening on multiple \acp{rv} at the same time, \ac{lci} is able to treat those \acp{rv} as a group.
For example, recall the employee example and assume we want to train multiple employees simultaneously as a training program is mostly offered not only for a single employee but for a group of employees.
Then, it is not necessary to split all trained employees into separate groups---it is sufficient to differentiate between trained employees and all remaining employees.
Formally, the interventions $do(R'_1 = r'_1, \dots, R'_k = r'_k)$ on an arbitrary set of \acp{rv} $\{R'_1, \dots, R'_k\}$ can thus efficiently be handled by splitting the \acp{pf} in $G$ such that all $R'_i$ that are represented by the same \ac{prv} $A$ and set to the same value $r'_i$ remain grouped, equal to splitting on constraints in \ac{lve}.
More specifically, \ac{lci} needs just a single split per group and thus avoids manipulating the parents of each individual \ac{rv} separately.
Furthermore, it is also possible to intervene on a \ac{prv} (instead of intervening on a \ac{rv}).
The semantics of an intervention on a \ac{prv} $A$ is given by $do(A = a) = do(R_1 = a, \dots, R_k = a)$ with $gr(A) = \{R_1, \dots, R_k\}$.
Again, \ac{lci} is able to treat all \acp{rv} represented by $A$ as a group and therefore is not required to split the group.
In contrast, in a propositional model, every object has to be treated individually and therefore the parents for each \ac{rv} need to be manipulated separately.

Next, we investigate the practical performance of \acp{pcfg} and, in particular, the \ac{lci} algorithm for the computation of interventional distributions.

\section{Experiments} \label{sec:lce_eval}
In this section, we evaluate the run times needed to compute the result of interventional queries in \aclp{bn}, directed \acp{fg}, and \acp{pcfg}.
For our experiments, we use a slightly modified version of the \ac{pcfg} given in \cref{fig:pcfg_example} which can directly be translated into a \acl{bn} without having to combine multiple parent factors into a single conditional probability table.
More specifically, to obtain the corresponding directed \ac{fg}, we simply ground the \ac{pcfg} and to obtain the equivalent \acl{bn}, we use the transformation from directed \ac{fg} to \acl{bn} given by \citet{Frey2003a}.
Note that the \ac{pcfg} used in our experiments to demonstrate the practical efficiency of lifted causal inference is rather small with four \acp{pf} and \acp{prv}, respectively, and the gain we obtain from lifted inference further increases with models consisting of more \acp{prv}.

\begin{figure}[t]
	\centering
	\begin{tikzpicture}[x=1pt,y=1pt]
		\definecolor{fillColor}{RGB}{255,255,255}
		\path[use as bounding box,fill=fillColor,fill opacity=0.00] (0,0) rectangle (289.08,130.09);
		\begin{scope}
		\path[clip] (  0.00,  0.00) rectangle (289.08,130.09);
		\definecolor{drawColor}{RGB}{255,255,255}
		\definecolor{fillColor}{RGB}{255,255,255}
		
		\path[draw=drawColor,line width= 0.6pt,line join=round,line cap=round,fill=fillColor] (  0.00,  0.00) rectangle (289.08,130.09);
		\end{scope}
		\begin{scope}
		\path[clip] ( 44.91, 30.69) rectangle (283.58,124.59);
		\definecolor{fillColor}{RGB}{255,255,255}
		
		\path[fill=fillColor] ( 44.91, 30.69) rectangle (283.58,124.59);
		\definecolor{drawColor}{RGB}{247,192,26}
		
		\path[draw=drawColor,line width= 0.6pt,line join=round] ( 58.95, 47.62) --
			( 82.35, 47.39) --
			(105.75, 47.88) --
			(129.15, 48.66) --
			(152.54, 49.68) --
			(175.94, 52.00) --
			(199.34, 57.04) --
			(222.74, 60.57) --
			(246.14, 62.72) --
			(269.54, 67.79);
		\definecolor{drawColor}{RGB}{78,155,133}
		
		\path[draw=drawColor,line width= 0.6pt,dash pattern=on 2pt off 2pt ,line join=round] ( 58.95, 34.95) --
			( 82.35, 36.16) --
			(105.75, 42.39) --
			(129.15, 50.19) --
			(152.54, 58.51) --
			(175.94, 68.66) --
			(199.34, 80.13) --
			(222.74, 92.69) --
			(246.14,106.03) --
			(269.54,120.32);
		\definecolor{drawColor}{RGB}{37,122,164}
		
		\path[draw=drawColor,line width= 0.6pt,dash pattern=on 4pt off 2pt ,line join=round] ( 58.95, 45.95) --
			( 82.35, 48.40) --
			(105.75, 51.82) --
			(129.15, 55.89) --
			(152.54, 61.28) --
			(175.94, 66.59) --
			(199.34, 73.10) --
			(222.74, 82.18) --
			(246.14, 94.26) --
			(269.54,107.99);
		\definecolor{fillColor}{RGB}{78,155,133}
		
		\path[fill=fillColor] ( 58.95, 38.01) --
			( 61.59, 33.43) --
			( 56.31, 33.43) --
			cycle;
		
		\path[fill=fillColor] ( 82.35, 39.21) --
			( 84.99, 34.63) --
			( 79.71, 34.63) --
			cycle;
		
		\path[fill=fillColor] (105.75, 45.44) --
			(108.39, 40.87) --
			(103.10, 40.87) --
			cycle;
		
		\path[fill=fillColor] (129.15, 53.24) --
			(131.79, 48.66) --
			(126.50, 48.66) --
			cycle;
		
		\path[fill=fillColor] (152.54, 61.56) --
			(155.19, 56.98) --
			(149.90, 56.98) --
			cycle;
		
		\path[fill=fillColor] (175.94, 71.71) --
			(178.59, 67.13) --
			(173.30, 67.13) --
			cycle;
		
		\path[fill=fillColor] (199.34, 83.18) --
			(201.99, 78.60) --
			(196.70, 78.60) --
			cycle;
		
		\path[fill=fillColor] (222.74, 95.74) --
			(225.38, 91.16) --
			(220.10, 91.16) --
			cycle;
		
		\path[fill=fillColor] (246.14,109.08) --
			(248.78,104.50) --
			(243.50,104.50) --
			cycle;
		
		\path[fill=fillColor] (269.54,123.37) --
			(272.18,118.79) --
			(266.90,118.79) --
			cycle;
		\definecolor{fillColor}{RGB}{37,122,164}
		
		\path[fill=fillColor] ( 56.99, 43.99) --
			( 60.91, 43.99) --
			( 60.91, 47.92) --
			( 56.99, 47.92) --
			cycle;
		\definecolor{fillColor}{RGB}{247,192,26}
		
		\path[fill=fillColor] ( 58.95, 47.62) circle (  1.96);
		\definecolor{fillColor}{RGB}{37,122,164}
		
		\path[fill=fillColor] ( 80.39, 46.44) --
			( 84.31, 46.44) --
			( 84.31, 50.37) --
			( 80.39, 50.37) --
			cycle;
		\definecolor{fillColor}{RGB}{247,192,26}
		
		\path[fill=fillColor] ( 82.35, 47.39) circle (  1.96);
		\definecolor{fillColor}{RGB}{37,122,164}
		
		\path[fill=fillColor] (103.78, 49.86) --
			(107.71, 49.86) --
			(107.71, 53.78) --
			(103.78, 53.78) --
			cycle;
		\definecolor{fillColor}{RGB}{247,192,26}
		
		\path[fill=fillColor] (105.75, 47.88) circle (  1.96);
		\definecolor{fillColor}{RGB}{37,122,164}
		
		\path[fill=fillColor] (127.18, 53.92) --
			(131.11, 53.92) --
			(131.11, 57.85) --
			(127.18, 57.85) --
			cycle;
		\definecolor{fillColor}{RGB}{247,192,26}
		
		\path[fill=fillColor] (129.15, 48.66) circle (  1.96);
		\definecolor{fillColor}{RGB}{37,122,164}
		
		\path[fill=fillColor] (150.58, 59.32) --
			(154.51, 59.32) --
			(154.51, 63.24) --
			(150.58, 63.24) --
			cycle;
		\definecolor{fillColor}{RGB}{247,192,26}
		
		\path[fill=fillColor] (152.54, 49.68) circle (  1.96);
		\definecolor{fillColor}{RGB}{37,122,164}
		
		\path[fill=fillColor] (173.98, 64.62) --
			(177.91, 64.62) --
			(177.91, 68.55) --
			(173.98, 68.55) --
			cycle;
		\definecolor{fillColor}{RGB}{247,192,26}
		
		\path[fill=fillColor] (175.94, 52.00) circle (  1.96);
		\definecolor{fillColor}{RGB}{37,122,164}
		
		\path[fill=fillColor] (197.38, 71.14) --
			(201.31, 71.14) --
			(201.31, 75.07) --
			(197.38, 75.07) --
			cycle;
		\definecolor{fillColor}{RGB}{247,192,26}
		
		\path[fill=fillColor] (199.34, 57.04) circle (  1.96);
		\definecolor{fillColor}{RGB}{37,122,164}
		
		\path[fill=fillColor] (220.78, 80.21) --
			(224.70, 80.21) --
			(224.70, 84.14) --
			(220.78, 84.14) --
			cycle;
		\definecolor{fillColor}{RGB}{247,192,26}
		
		\path[fill=fillColor] (222.74, 60.57) circle (  1.96);
		\definecolor{fillColor}{RGB}{37,122,164}
		
		\path[fill=fillColor] (244.18, 92.30) --
			(248.10, 92.30) --
			(248.10, 96.22) --
			(244.18, 96.22) --
			cycle;
		\definecolor{fillColor}{RGB}{247,192,26}
		
		\path[fill=fillColor] (246.14, 62.72) circle (  1.96);
		\definecolor{fillColor}{RGB}{37,122,164}
		
		\path[fill=fillColor] (267.58,106.03) --
			(271.50,106.03) --
			(271.50,109.96) --
			(267.58,109.96) --
			cycle;
		\definecolor{fillColor}{RGB}{247,192,26}
		
		\path[fill=fillColor] (269.54, 67.79) circle (  1.96);
		\end{scope}
		\begin{scope}
		\path[clip] (  0.00,  0.00) rectangle (289.08,130.09);
		\definecolor{drawColor}{RGB}{0,0,0}
		
		\path[draw=drawColor,line width= 0.6pt,line join=round] ( 44.91, 30.69) --
			( 44.91,124.59);
		
		\path[draw=drawColor,line width= 0.6pt,line join=round] ( 46.33,122.12) --
			( 44.91,124.59) --
			( 43.49,122.12);
		\end{scope}
		\begin{scope}
		\path[clip] (  0.00,  0.00) rectangle (289.08,130.09);
		\definecolor{drawColor}{gray}{0.30}
		
		\node[text=drawColor,anchor=base east,inner sep=0pt, outer sep=0pt, scale=  0.88] at ( 39.96, 45.71) {10};
		
		\node[text=drawColor,anchor=base east,inner sep=0pt, outer sep=0pt, scale=  0.88] at ( 39.96, 69.44) {100};
		
		\node[text=drawColor,anchor=base east,inner sep=0pt, outer sep=0pt, scale=  0.88] at ( 39.96, 93.17) {1000};
		
		\node[text=drawColor,anchor=base east,inner sep=0pt, outer sep=0pt, scale=  0.88] at ( 39.96,116.89) {10000};
		\end{scope}
		\begin{scope}
		\path[clip] (  0.00,  0.00) rectangle (289.08,130.09);
		\definecolor{drawColor}{gray}{0.20}
		
		\path[draw=drawColor,line width= 0.6pt,line join=round] ( 42.16, 48.74) --
			( 44.91, 48.74);
		
		\path[draw=drawColor,line width= 0.6pt,line join=round] ( 42.16, 72.47) --
			( 44.91, 72.47);
		
		\path[draw=drawColor,line width= 0.6pt,line join=round] ( 42.16, 96.20) --
			( 44.91, 96.20);
		
		\path[draw=drawColor,line width= 0.6pt,line join=round] ( 42.16,119.92) --
			( 44.91,119.92);
		\end{scope}
		\begin{scope}
		\path[clip] (  0.00,  0.00) rectangle (289.08,130.09);
		\definecolor{drawColor}{RGB}{0,0,0}
		
		\path[draw=drawColor,line width= 0.6pt,line join=round] ( 44.91, 30.69) --
			(283.58, 30.69);
		
		\path[draw=drawColor,line width= 0.6pt,line join=round] (281.12, 29.26) --
			(283.58, 30.69) --
			(281.12, 32.11);
		\end{scope}
		\begin{scope}
		\path[clip] (  0.00,  0.00) rectangle (289.08,130.09);
		\definecolor{drawColor}{gray}{0.20}
		
		\path[draw=drawColor,line width= 0.6pt,line join=round] ( 58.95, 27.94) --
			( 58.95, 30.69);
		
		\path[draw=drawColor,line width= 0.6pt,line join=round] ( 82.35, 27.94) --
			( 82.35, 30.69);
		
		\path[draw=drawColor,line width= 0.6pt,line join=round] (105.75, 27.94) --
			(105.75, 30.69);
		
		\path[draw=drawColor,line width= 0.6pt,line join=round] (129.15, 27.94) --
			(129.15, 30.69);
		
		\path[draw=drawColor,line width= 0.6pt,line join=round] (152.54, 27.94) --
			(152.54, 30.69);
		
		\path[draw=drawColor,line width= 0.6pt,line join=round] (175.94, 27.94) --
			(175.94, 30.69);
		
		\path[draw=drawColor,line width= 0.6pt,line join=round] (199.34, 27.94) --
			(199.34, 30.69);
		
		\path[draw=drawColor,line width= 0.6pt,line join=round] (222.74, 27.94) --
			(222.74, 30.69);
		
		\path[draw=drawColor,line width= 0.6pt,line join=round] (246.14, 27.94) --
			(246.14, 30.69);
		
		\path[draw=drawColor,line width= 0.6pt,line join=round] (269.54, 27.94) --
			(269.54, 30.69);
		\end{scope}
		\begin{scope}
		\path[clip] (  0.00,  0.00) rectangle (289.08,130.09);
		\definecolor{drawColor}{gray}{0.30}
		
		\node[text=drawColor,anchor=base,inner sep=0pt, outer sep=0pt, scale=  0.88] at ( 58.95, 19.68) {8};
		
		\node[text=drawColor,anchor=base,inner sep=0pt, outer sep=0pt, scale=  0.88] at ( 82.35, 19.68) {16};
		
		\node[text=drawColor,anchor=base,inner sep=0pt, outer sep=0pt, scale=  0.88] at (105.75, 19.68) {32};
		
		\node[text=drawColor,anchor=base,inner sep=0pt, outer sep=0pt, scale=  0.88] at (129.15, 19.68) {64};
		
		\node[text=drawColor,anchor=base,inner sep=0pt, outer sep=0pt, scale=  0.88] at (152.54, 19.68) {128};
		
		\node[text=drawColor,anchor=base,inner sep=0pt, outer sep=0pt, scale=  0.88] at (175.94, 19.68) {256};
		
		\node[text=drawColor,anchor=base,inner sep=0pt, outer sep=0pt, scale=  0.88] at (199.34, 19.68) {512};
		
		\node[text=drawColor,anchor=base,inner sep=0pt, outer sep=0pt, scale=  0.88] at (222.74, 19.68) {1024};
		
		\node[text=drawColor,anchor=base,inner sep=0pt, outer sep=0pt, scale=  0.88] at (246.14, 19.68) {2048};
		
		\node[text=drawColor,anchor=base,inner sep=0pt, outer sep=0pt, scale=  0.88] at (269.54, 19.68) {4096};
		\end{scope}
		\begin{scope}
		\path[clip] (  0.00,  0.00) rectangle (289.08,130.09);
		\definecolor{drawColor}{RGB}{0,0,0}
		
		\node[text=drawColor,anchor=base,inner sep=0pt, outer sep=0pt, scale=  1.10] at (164.24,  7.64) {$d$};
		\end{scope}
		\begin{scope}
		\path[clip] (  0.00,  0.00) rectangle (289.08,130.09);
		\definecolor{drawColor}{RGB}{0,0,0}
		
		\node[text=drawColor,rotate= 90.00,anchor=base,inner sep=0pt, outer sep=0pt, scale=  1.10] at ( 13.08, 77.64) {time (ms)};
		\end{scope}
		\begin{scope}
		\path[clip] (  0.00,  0.00) rectangle (289.08,130.09);
		
		\path[] ( 42.12, 73.93) rectangle (200.45,128.29);
		\end{scope}
		\begin{scope}
		\path[clip] (  0.00,  0.00) rectangle (289.08,130.09);
		\definecolor{drawColor}{RGB}{247,192,26}
		
		\path[draw=drawColor,line width= 0.6pt,line join=round] ( 49.06,115.56) -- ( 60.63,115.56);
		\end{scope}
		\begin{scope}
		\path[clip] (  0.00,  0.00) rectangle (289.08,130.09);
		\definecolor{fillColor}{RGB}{247,192,26}
		
		\path[fill=fillColor] ( 54.84,115.56) circle (  1.96);
		\end{scope}
		\begin{scope}
		\path[clip] (  0.00,  0.00) rectangle (289.08,130.09);
		\definecolor{drawColor}{RGB}{78,155,133}
		
		\path[draw=drawColor,line width= 0.6pt,dash pattern=on 2pt off 2pt ,line join=round] ( 49.06,101.11) -- ( 60.63,101.11);
		\end{scope}
		\begin{scope}
		\path[clip] (  0.00,  0.00) rectangle (289.08,130.09);
		\definecolor{fillColor}{RGB}{78,155,133}
		
		\path[fill=fillColor] ( 54.84,104.16) --
			( 57.49, 99.59) --
			( 52.20, 99.59) --
			cycle;
		\end{scope}
		\begin{scope}
		\path[clip] (  0.00,  0.00) rectangle (289.08,130.09);
		\definecolor{drawColor}{RGB}{37,122,164}
		
		\path[draw=drawColor,line width= 0.6pt,dash pattern=on 4pt off 2pt ,line join=round] ( 49.06, 86.66) -- ( 60.63, 86.66);
		\end{scope}
		\begin{scope}
		\path[clip] (  0.00,  0.00) rectangle (289.08,130.09);
		\definecolor{fillColor}{RGB}{37,122,164}
		
		\path[fill=fillColor] ( 52.88, 84.69) --
			( 56.81, 84.69) --
			( 56.81, 88.62) --
			( 52.88, 88.62) --
			cycle;
		\end{scope}
		\begin{scope}
		\path[clip] (  0.00,  0.00) rectangle (289.08,130.09);
		\definecolor{drawColor}{RGB}{0,0,0}
		
		\node[text=drawColor,anchor=base west,inner sep=0pt, outer sep=0pt, scale=  0.80] at ( 67.57,112.81) {Lifted Variable Elimination (PCFG)};
		\end{scope}
		\begin{scope}
		\path[clip] (  0.00,  0.00) rectangle (289.08,130.09);
		\definecolor{drawColor}{RGB}{0,0,0}
		
		\node[text=drawColor,anchor=base west,inner sep=0pt, outer sep=0pt, scale=  0.80] at ( 67.57, 98.36) {Variable Elimination (BN)};
		\end{scope}
		\begin{scope}
		\path[clip] (  0.00,  0.00) rectangle (289.08,130.09);
		\definecolor{drawColor}{RGB}{0,0,0}
		
		\node[text=drawColor,anchor=base west,inner sep=0pt, outer sep=0pt, scale=  0.80] at ( 67.57, 83.90) {Variable Elimination (FG)};
		\end{scope}
	\end{tikzpicture}	
	\caption{A comparison of the run times required to compute interventional distributions on different graphical models encoding equivalent full joint probability distributions.}
	\label{fig:plot_eval}
\end{figure}
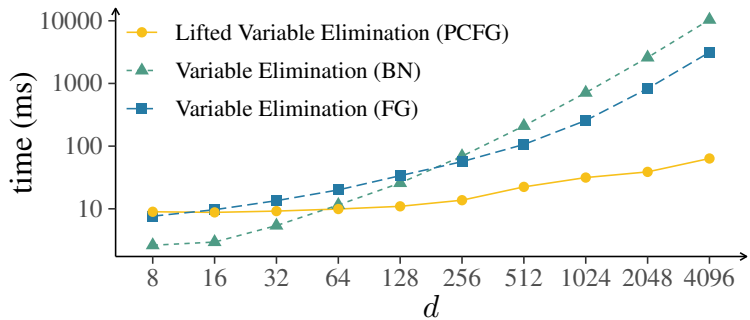

We test the required run time for each of the three graphical models on different graph sizes by setting the domain size of the employees to $d = 8, 16, 32, \dots, 4096$ and having a single training program for each choice of $d$ (i.e., $\abs{\mathcal D(E)} = d$ and $\abs{\mathcal D(T)} = 1$).
\Cref{fig:plot_eval} shows the run times needed to compute an interventional distribution for a single intervention in the modified graph when running \acl{ve} on the directed \ac{fg}, \acl{ve} on the \acl{bn}, and \ac{lve} on the \ac{pcfg}.
The \acl{ve} algorithm is the propositional counterpart of \ac{lve} and operates on a propositional (ground) model, such as a \acl{bn} or an \ac{fg}.
Consequently, \acl{ve} considers every object represented by a \ac{rv} (e.g., every employee) individually for computations, independent of whether there are objects behaving exactly the same.
In contrast, \ac{lve} treats identically behaving objects as a group by using a representative for computations instead of considering each of those objects separately.
The results emphasise that the \ac{lci} algorithm, which internally exploits \ac{lve}, overcomes scalability issues for large domain sizes as the run time of \ac{lve}, in contrast to the run times of \acl{ve} on the \acl{bn} and the directed \ac{fg}, does not exponentially increase with $d$ (y-axis is log-scaled).
To conclude, \acp{pcfg} not only provide us with expressive probabilistic graphical models for relational domains but also enable us to drastically speed up causal inference by reasoning over sets of indistinguishable objects.

\section{Conclusion} \label{sec:lce_conclusion}
We introduce \acp{pcfg} to combine lifted probabilistic inference in relational domains with causal inference, thereby allowing for lifted causal inference.
To leverage the power of lifted inference for causal effect computation, we present the \ac{lci} algorithm, which operates on a lifted level and thus allows us to drastically speed up causal inference. 
\Ac{lci} is a simple, yet effective algorithm to compute the effect of (multiple simultaneous) interventions, and builds on the well-founded \ac{lve} algorithm, thereby allowing \ac{lci} to be plugged into parameterised decision models~\citep{Gehrke2019b} to compute the maximum expected utility in accordance with \citet{Pearl2009a}.

\Acp{pcfg} open up interesting directions for future work.
A basic problem is to learn a \ac{pcfg} directly from a relational database.
Following up on learning \acp{pcfg} from data, another constitutive problem for future research is to relax the assumption of having a fully directed \ac{pcfg} at hand, i.e., to allow \acp{pcfg} to contain both directed and undirected edges at the same time and investigate the implications for answering causal queries.

\acks{This work is partially funded by the BMBF project AnoMed 16KISA057 and 16KISA050K, and is also partially supported by the Medical Cause and Effects Analysis (MCEA) project and partially funded by the Deutsche Forschungsgemeinschaft (DFG, German Research Foundation) under Germany's Excellence Strategy -- EXC 2176 \enquote{Understanding Written Artefacts: Material, Interaction and Transmission in Manuscript Cultures}, project no. 390893796.}

\bibliography{references.bib}
\end{document}